# Discovery of Important Crossroads in Road Network using Massive Taxi Trajectories

Ming Xu[1], Jianping Wu[2], Yiman Du[2,*], Haohan Wang[3], Geqi Qi[2], Kezhen Hu[2], Yunpeng Xiao[4]
[1]School of Computer Science, Beijing University of Posts and Telecommunications, Beijing, China
[2]Department of Civil Engineering, Tsinghua University, Beijing, China
[3]School of Computer Science, Carnegie Mellon University, Pittsburgh PA USA
[4]School of Software, Chongqing University of Posts and Telecommunications, Chongqing, China
{xum, xiaoyp}@bupt.edu.cn, {2855175, 153789104}@qq.com

## ABSTRACT
A major problem in road network analysis is discovery of important crossroads, which can provide useful information for transport planning. However, none of existing approaches addresses the problem of identifying network-wide important crossroads in real road network. In this paper, we propose a novel data-driven based approach named CRRank to rank important crossroads. Our key innovation is that we model the trip network reflecting real travel demands with a tripartite graph, instead of solely analysis on the topology of road network. To compute the importance scores of crossroads accurately, we propose a HITS-like ranking algorithm, in which a procedure of score propagation on our tripartite graph is performed. We conduct experiments on CRRank using a real-world dataset of taxi trajectories. Experiments verify the utility of CRRank.

## Categories and Subject Descriptors
D.3.3 [**Analysis of Algorithms and Problem Complexity**]: Nonnumerical Algorithms and Problems – *Pattern matching.*

## General Terms
Algorithms

## Keywords
Tripartite graph, Ranking algorithm, Road network, Taxi trajectories,

## 1. INTRODUCTION
So far, urban transportation system is one of the largest and most complex systems closely related with people. In such a system, road network plays a key role, for it carries all the daily traffic flow. The understanding and interpretation of reliability of road network is a crucial component to improve transportation efficiency, which attracts a large number of researchers. In general, road network is abstracted as graph structure. In this structure, nodes represent crossroads, and edges represent road segment linking two neighborhood crossroads.

De Montis et al. [1] find that an obvious scale free characteristic is showed in the road network, and the relationship between nodes and traffics follows a power law distribution, which means, the main traffic volume is carried on a small number of nodes. Lämmer et al. [2] find that road network shows a phenomenon of cascading failures. In other words, when a few crucial nodes fail, some other nodes would also fail because of the connection between the nodes, and this situation can even lead to the collapse of the entire road network. Therefore, how to identify such a few significant nodes accurately, and then focus on them for control strategies, is an effective approach for reducing congestion and improving efficiency of road network.

In this paper, we aim to discover the important nodes in the road network using a novel data-driven framework named CRRank. In out method, the trip network which consists of dynamic transport information like traffic flow, Origin-Destination (OD) and paths of trips is considered instead of static road network. Firstly, we extract trip information from massive daily taxi trajectories and road network, and then we model the trip network using a tripartite graph. Inspired by HITS algorithm [9], which ranks web documents based on link information among nodes in a hyperlink graph, we propose a mutual reinforcing algorithm, which performs an iterative procedure of score propagation over our tripartite graph to rank the network-wide important crossroads.

The remainder of the paper is organized as follows. In Section 2, related studies of discovery of important nodes are presented. In Section 3, we briefly discuss original intention behind our model. Then, proposed framework and related definitions are introduced in this section. In Section 4, basic definition of tripartite graph, and how it can be used in the trip network modeling are described. Also, an algorithm of score propagation on tripartite graph is presented here. Experiments on validity of CRRank algorithm are presented in Section 5. Section 6 concludes the paper with a summary.

## 2. RELATED WORK
Many previous methods of searching for important nodes are widely studied in complex network and graph mining. In these methods, some metrics are proposed to accomplish this task. Zhao et al. [3] find that degree distribution of a node is closely related to its importance. In detail, hubs or high connectivity nodes at the tail of the power-law degree distribution are known as important ones and play key roles in maintaining the functions of networks. Betweenness centrality is also used to find crucial central nodes [4], and Wang et al. [5] introduce betweenness centrality to analyze air transport network. A method to measure the node importance by combining degree distribution and clustering coefficient information is presented in [6]. Unfortunately, these metrics cannot be directly applied to road network, since each node in road networks exhibits a very limited range of degrees, and almost all the nodes degrees are similar. Moreover, these metrics only reflect characteristics of topology of the road network, some dynamic elements, such as traffic flows and OD information, are lost by utilizing them straightforwardly.

---

* *correspondence author of this paper*

Another approach is called system analysis approach. The basic idea of such approach is evaluating cascading failure after removing each node. Combining such idea and characteristics of transport system, Wu et al. [7] present a cascading failure transportation network evaluation model, which can reveal deep knowledge of network. However, such theory-driven model has to be analyzed on a simulated network instead of real road network, while the underlying logic of the real road network can hardly be grasped for simulation. In addition, some data-driven approaches depending on the calculation of the eigenvectors are successfully used to find important web or information on internet. Google PageRank [8] and HITS are famous examples in this category. This work is inspired by this kind of methods. A major difference of CRRank from approaches previously mentioned is that our data-driven algorithm is based on the tripartite graph which incorporates traffic, travel demands, empirical knowledge of drivers and topology of road network, and CRRank can identify network-wide important crossroads effectively.

In addition, the flourishing of sensing technologies and large-scale computing infrastructures have produced a variety of big and heterogeneous data in urban spaces, e.g., human mobility, vehicle trajectories, air quality, traffic patterns, and geographical data [10]. These data brings out a variety of novel data mining methods for urban planning and road network analysis. A LDA-based inference model combining road network data, points of interests, and a large number of taxi trips to infer the functional regions in a city is proposed in [11]. Zheng et al. [12] discover the bottleneck of road network using the corresponding taxi trajectories. Chawla et al.[13] present a PCA-based algorithm to identify link anomalies, and then $L_1$ optimization techniques is utilized to infer the traffic flows that lead to a traffic anomaly. Liu et al. [14] use frequent subgraph mining to discover anomalous links for each time interval, and then form outlier trees to reveal the potential flaws in the design of existing traffic networks. Being similar with these studies, this paper also addresses a problem of transport planning using a large number of taxi trajectories.

## 3. OVERVIEW
### 3.1 Motivation
To illustrate our idea, consider a simplified road network as an example in Figure 1. Each circle represents a crossroad, and each directed edge linking two neighbor circles represents a road segment. These road segments partition the map into some rectangle regions. Our goal is to compare the importance of two crossroads $g$ and $j$. A simple method is to measure the betweenness and traffic flow of each crossroad. Both of them have the same betweenness due to the symmetrical topology. Therefore, traffic flow becomes the only decisive factor. Intuitively, the crossroad which carries the bigger traffic flow may be more important. However, suppose that traffic of the two crossroads are equal, both of them seem to be equally important, unless we can get additional information. Now suppose that traffic flow of $j$ is mainly caused by trips between two neighbor regions $G$ and $H$, while traffic flows of $g$ is generated by a wider range of OD pairs. Obviously, crossroad $g$, which acts as a traffic hub, is more important than crossroad $j$. Consequently, we can conclude that the importance of a crossroad is related to OD distribution of trips as well as traffic flow. In fact, transportation network includes not only road network, but also the trip network which reflects travel demands. The latter can provide more valuable information for locating important crossroads.

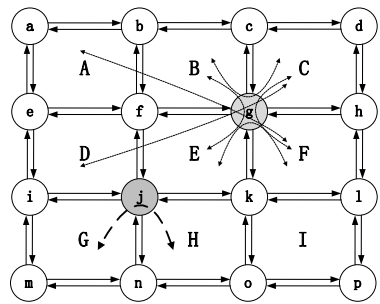

**Figure 1. Insert caption to place caption below.**

### 3.2 Framework of our method
Before giving our approach, we first introduce some terms used in this paper.

*Definition 1 (Road Network)*: A road network is a directed weighted graph $G(V,E)$, where $V$ is set of nodes representing crossroads or end points of road segments, and $E$ is set of weighted edges representing road segments linking two crossroads, a node $v$ is uniquely identified with an id $v.id$ and an edge $e$ is associated with an id $e.id$. $e.level$ indicates the level of $e$. In detail, if $e$ is an expressway, freeway or arterial road, $e.level$ is equal to 1, if $e$ is a sub-arterial road, $e.level$ is equal to 2, if $e$ is a bypass, $e.level$ is equal to 3.

*Definition 2 (Region)*: A city map can be partitioned into some non-overlapping regions by high level road segments in road network. The set of these regions is represented by $\Omega = \{g_i\}_{i=1}^{|\Omega|}$, where $|\Omega|$ is total number of Regions. A region $g$ is associated with an uniquely identifier $g.id$, some low level road segments can be included within a region，this method of map partition can preserve semantic information of regions，e.g. school, park, business areas or resident area etc.

*Definition 3 (Transition)*: A transition $t$ is represented by a triple $<o,d,P_t>$, where $o$ represents origin region, $d$ represents destination region, $o,d \in \Omega$, $P_t = \{p_i\}_{i=1}^{|P_t|}$ is the set of paths for transition $t$, $|P_t|$ is the number of paths of $t$.

*Definition 4 (path)*: A path $p$ is a crossroads sequence including all crossroads traversed by a trip $l$, which is an effective trajectory corresponding to a transition $t$. $p$ can be denoted by $\{v_i\}_{i=1}^{|p|}$, where $v_i$ is a crossroad and $|p|$ is the number of crossroads in path $p$. The length of a path $p$ is the number of crossroads in $p$.

We model trip network using a tripartite graph for characterizing the relationship among transitions, paths and crossroads. Figure 2 shows the framework of our method, it consists of two major parts: information extraction and trip network analysis.

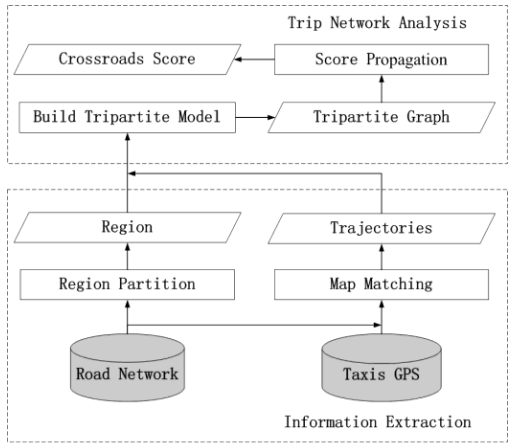

Figure 2. The framework of CRRank

Information Extraction: due to the noises of the GPS records, the GPS points often deviate from the actual road segments. In this phase, we need to map each GPS reading of each taxi trajectory onto a road segment using a map matching algorithm [15]. Each mapped GPS reading is represented by a triple $<l.id, e.id, tm>$, where $l.id$ is the identifier of a trip, $e.id$ is the identifier of a road segment and $tm$ is sampling time. In addition, we partition the urban area of map into disjoint regions with high level road segments using area segmentation algorithm [16].

Trip Network Analysis: In this phase, we design a tripartite graph to model trip network using data handled by information extraction phase. In order to construct the tripartite graph, we need to discover the relationship among transitions, paths and crossroads. A transition can be obtained according to regions in which the origin road and the destination road of a trajectory are. An example is given in Figure 3. The origin of the transition is $g_1$ and destination is $g_3$. This transition contains three paths $p_1$, $p_2$ and $p_3$. However, if the endpoint of the trajectory is located on the major road, which is used to partition regions, it is difficult to determine the OD of this transition. An example is given in Figure 4, an endpoint $o_1$ of this trajectory is at the junction of two neighbor regions $g_1$ and $g_5$, and the other endpoint $o_2$ is at the junction of $g_4$ and $g_7$. In this case, we allocate the origin and destination to the right region due to the Chinese traffic rule which requires driving on the right side. To explain it concretely, we assume that the transition is from $o_1$ to $o_2$, the origin is $g_5$ and destination is $g_7$. If the transition is in the opposite direction, the origin is $g_4$ and destination is $g_1$. Although it is not always consistent with the facts, this allocation is feasible and reasonable. Finally, the tripartite graph is input to sore propagation component to rank important crossroads. Apart from ranking important crossroads, there are several more advantages using a tripartite graph to model the trip network, such as locating the popular paths and identifying the transitions, which have a large impact on the road network.

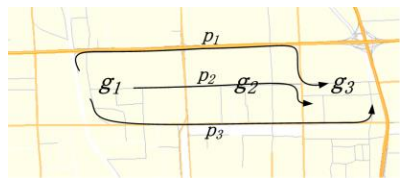

Figure 3. A transition that contains three paths.

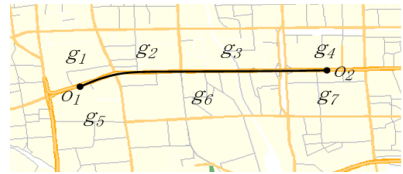

Figure 4. A transition whose OD are on the major road.

## 4. CRRANK ALGORITHM

### 4.1 Tripartite Graph

A tripartite graph is a graph whose nodes can be partitioned into three disjoint sets so that no two nodes within the same set are directly connected. The relationships encoded by the edges between two sets of nodes can be summarized into a pair of adjacency matrices In this paper, we model trip network by a tripartite graph, as Figure 5 presented, which can be represented by $G'(T \cup P \cup V, W \cup U)$ formally, where $T$, $P$ and $V$ represents the sets of nodes corresponding to transitions, paths, and crossroads respectively, $W$ denotes the adjacency matrix for the transition-path edges, the entry $w_{m,k}$ of $W$ is the number of trips corresponding transition $m$, which selects path $k$. $U$ denotes the corresponding matrix for path-crossroad edges. If path $k$ contains crossroad $n$, $u_{k,n}$ equals 1, otherwise, $u_{k,n}$ equals 0.

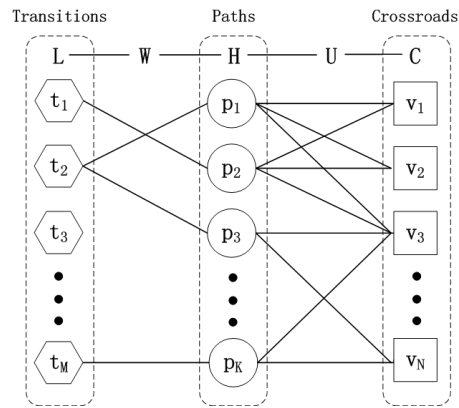

Figure 5. The tripartite graph of trip network

The dynamic transportation information on the road network is modeled using a tripartite graph. Next, we present a score propagation algorithm borrowing HITS to rank the importance of crossroads based on the link structure of the graph.

### 4.2 Score Propagation Algorithm

Mutually reinforcing relationships among linked objects in graphs are ubiquitous and considered as basis of many ranking algorithm,

e.g., in HITS algorithm, each webpage has both a hub score and an authority score. The intuition is that a good authority is pointed to many good hubs and a good hub points to many good authorities. The hub and authority of each webpage will converge to reasonable values via an iterative calculation, which is similar to the idea of our algorithm.

In order to perform score propagation over our tripartite graph, each node is assigned a meaningful score, which is updated during each iteration. We use a vector $L$ named transition load to represent the scores of transition set $T$. In fact, road network is limited as an important transportation resource. Once a trip of any transition is generated, the load of road network is heightened. The load caused by different transitions would be different. Let $H$ be the score vector, which represents the popularity of paths, and $C$ be the score vector, which represents the importance of crossroads. In order to integrate more priori knowledge about travel demand, traffic flow and topology of road network into CRRank, we define three profile vectors $L^{(0)}$, $H^{(0)}$ and $C^{(0)}$, which are initial score lists of transition loads, path popularities and crossroad importance, these vectors can be defined as

$$L^{(0)} = \left[ \frac{\tau_1}{\sum_m \tau_m} \quad \frac{\tau_2}{\sum_m \tau_m} \quad \cdots \quad \frac{\tau_M}{\sum_m \tau_m} \right]^T$$

$$H^{(0)} = \left[ \frac{\eta_1}{\sum_k \eta_k} \quad \frac{\eta_2}{\sum_k \eta_k} \quad \cdots \quad \frac{\eta_K}{\sum_k \eta_k} \right]^T$$

$$C^{(0)} = \left[ \frac{\sum_{d_1} \gamma_1^{d_1}}{\sum_n \sum_{d_n} \gamma_n^{d_n}} \quad \frac{\sum_{d_2} \gamma_2^{d_2}}{\sum_n \sum_{d_n} \gamma_n^{d_n}} \quad \cdots \quad \frac{\sum_{d_N} \gamma_N^{d_N}}{\sum_n \sum_{d_n} \gamma_n^{d_n}} \right]^T$$

Where $\tau_m$ denotes the number of trips corresponding to transition $m$, $\eta_k$ is the number of trips selecting path $k$, $\gamma_n^{d_n}$ is level score of $d_n^{th}$ edge of crossroad $n$, which is defined as

$$\gamma_n^{d_n} = \begin{cases} 1+\lambda & \text{if } e.level = 1 \\ 1 & \text{if } e.level = 2 \\ 1-\lambda & \text{if } e.level = 3 \end{cases}$$

Here, value of $\lambda$ is set to 0.2. These definitions are intuitive, the transition with large traffic flow is perhaps a heavy load to road network, and the path selected by many trips, has a high initial popularity score, and the crossroad linking high level roads may be important. The final scores of all nodes can be synchronously calculated with the weight matrix via an iterative mode. The weighted matrixes between $T$ and $P$ can be obtained by transforming from adjacency matrix using column normalization, which respectively represented by $X_{TP}$ and $X_{PT}$, are presented as

$$X_{TP} = \left[ \frac{w_{k,m}}{\sum_{i \in P_k} w_{i,m}} \right]_{K \times M}$$

$$X_{PT} = X_{TP}^T$$

And weighted matrixes between $P$ and $V$ are calculated by

$$Y_{PV} = \left[ u_{n,k} c_n^{(0)} \right]_{N \times K}$$

$$Y_{VP} = \left[ \frac{u_{n,k} h_k^{(0)}}{\sum_k u_{n,k} h_k^{(0)}} \right]_{N \times K}^T$$

Here, $c_n^{(0)}$ is the $n^{th}$ entry of vector $C^{(0)}$ and $h_k^{(0)}$ is the $k^{th}$ entry of vector $H^{(0)}$. The score calculation on our tripartite graph consists of a forward phase and a reverse phase. In forward phase, scores propagation begins from transition nodes, and the transition load scores are transformed into the popularity scores of their neighboring path nodes through the weighted matrix $X_{TP}$. Then importance score can be yielded using weighted matrix $Y_{PV}$ via a similar procedure. $H$ and $C$ can be updated successively as

$$H = \alpha X_{TP} L + (1-\alpha) H^{(0)}$$

$$C = \alpha Y_{PV} H + (1-\alpha) C^{(0)}$$

In reverse phase, the scores spread in the opposite direction. $H$ and $L$ can be updated successively as

$$H = \alpha Y_{VP} C + (1-\alpha) H^{(0)}$$

$$L = \alpha X_{PT} H + (1-\alpha) L^{(0)}$$

The forward phase and the reverse phase are applied in an alternating fashion until convergence. To avoid the scores for unlimited increasing, $H$, $C$ and $L$ needs to be normalized so that $\sum_k h_k = 1$, $\sum_n c_n = 1$ and $\sum_m l_m = 1$ after every iteration. The meaning behind our score propagation algorithm is the mutual reinforcement to boost co-linked entities on the tripartite graph. In detail, a popular path is likely selected by a heavy load transition, and a heavy load transition probably has a small number of or only one path to select; similarly, an important crossroad is likely included in many popular paths and in all probability, a popular path consists of many important crossroads. CRRank algorithm has some charming advantages: a variety of information such as traffic, travel demand, and topology of road network, and empirical knowledge of drivers routing are incorporated into a tripartite graph model, and the relevance score of each crossroad can be computed from a network-wide perspective.

## 5. EXPERIMENTS
### 5.1 Experimental set up and datasets
Road network: The road network of Beijing contains 13722 crossroads and 25178 road segments. The map is segmented into 478 regions using high level roads. As illustrate in Figure 6, the number $f_1(k)$ of trips of $k^{th}$ most frequently visited crossroads follows an exponential distribution:

$$f_1(k) \sim -7.242 \times 10^{-12} \times e^{5.46k} + 6.265 \times e^{-0.2202k}$$

It indicates that most of trips of taxis are carried on several crossroads, which may be candidates of important crossroads.

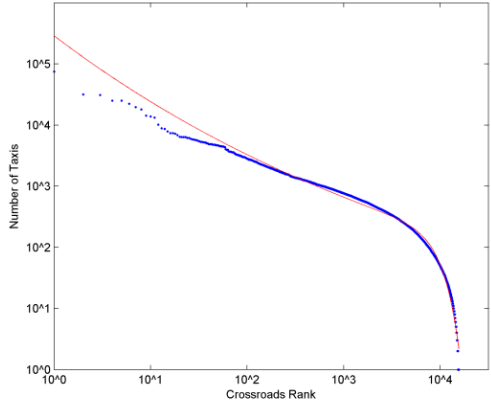

Figure 6. The number of trips of kth most frequent passed crossroad follows an exponential distribution

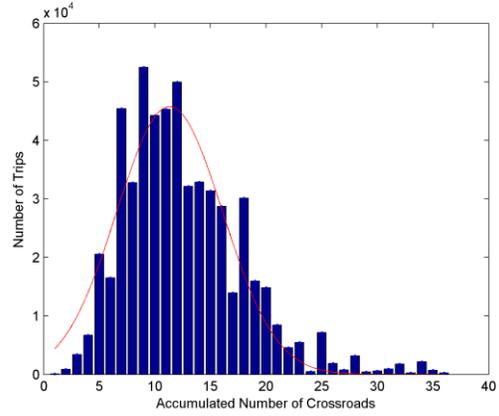

Figure 7. The number of trips which pass different number of crossroads follows a Gauss distribution

Taxi trajectories: we use a trajectory dataset generated by approximately 10000 taxicabs in Beijing over a period of one month (October in 2012). The sampling interval of the dataset is between 30s and 60s. Considering the morning and evening traffic peaks of road network, we select the trajectories from the dataset for our experiments, whose sampling timestamp ranges from 7:30 to 10:00 and from 17:00 to 19:30. As a result, we extract the 565,713 effective trips (the taxi was carrying a passenger). We measure the number of crossroads traversed by each trip. As illustrated in Figure 7. The number $f_2(q)$ of trips which pass number $q$ crossroads can be fitted using a Gauss distribution:

$$f_2(q) \sim 4.576 \times 10^{0.04} \times e^{-\left(\frac{q-11.32}{6.739}\right)^2}$$

We can find that the number of crossroads traversed by each trip mainly range from seven to eighteen. Due to noises of trajectories and mismatches of map matching, we remove the ineffective transitions which have less than 3 trips. As a result, we obtain 4628 transitions, 11297 paths. Our tripartite graph is built depending on this information.

## 5.2 Result

We compare our CRRank algorithm with a baseline: a naïve algorithm proposed in [17], it computes importance scores of crossroads according to the traffic flow of the crossroad and topology. The topology information contains degree of crossroads and level of road segments linking the crossroad. Figure 8 and 9 present the visualized results of ranking important crossroads of CRRank and baseline algorithm. The crossroads marked with the biggest red points are the top 5 most important nodes. The blue points represent 6[th] to 25[th] most important crossroads; the orange points represent the left of top 100 most important crossroads. The rest of crossroads are marked with yellow points. The parts of ranking results are also listed in Table 1 and 2.

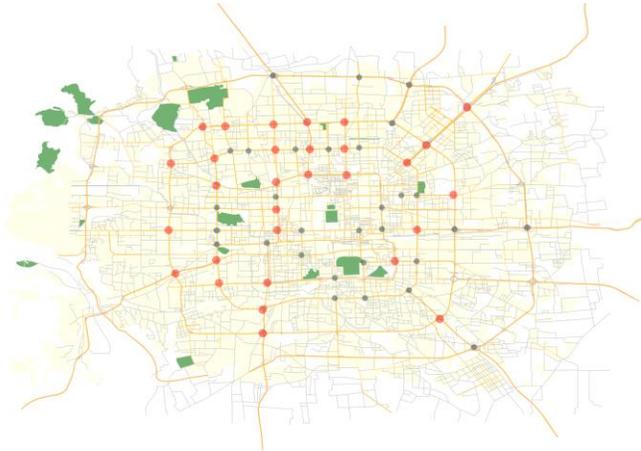

Figure 8. The important crossroads in the results of CRRank.

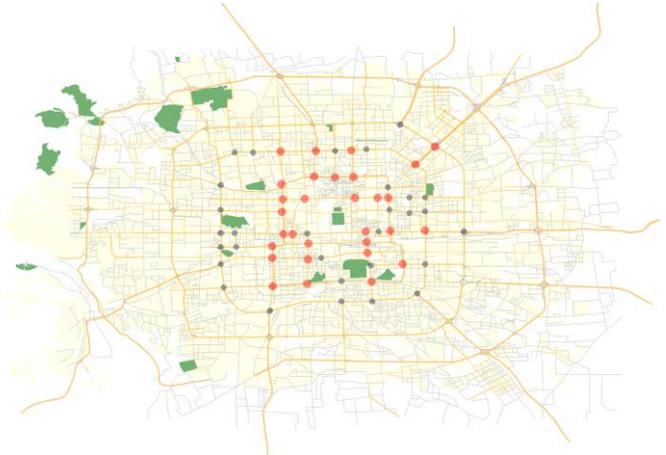

Figure 9. The important crossroads in the results of Baseline algorithm.

We observe that, the 1[st] and 2[nd] entries of results of two methods are the same. However, the important crossroads discovered by CRRank are mainly located on the ring roads with a wider range, while baseline algorithm is inclined to recognize the most important crossroads within 2rd road. The underlying reason for the interesting results is that, some crossroads, which are within 2rd ring road, have a large traffic flow and seem to be transport hubs, but the number of OD pairs which they connect is not so high. In many cases, they are the origins or destinations. For example, Crossroad "Xidan" near the largest Shopping Centre in Beijing and crossroad "Beijing Station Street-Jianguomenwai Street" near the Beijing Station have high ranks in the results of baseline algorithm. But in daily trips, drivers often prefer to choose ring roads to avoid these crossroads. In reality, the ring roads carry more trips generated by different transitions. To guarantee the smooth of the ring roads, many overpasses are built instead of traditional crossroads. The reduction of crossroads on ring roads leads to the fact that more transitions select ring roads instead of the shortest route, consequently, the importance of the remaining crossroads on the ring roads are boosting. It can be inferred that, once an unexpected accident occurs on the ring roads, more transitions will be delayed in various degrees.

**Table 1. Important crossroads rank using CRRank**

| Rank | Name or Position | Scope |
|---|---|---|
| 1 | Deshenmen Bridge | $2^{nd}$ ring road |
| 2 | Xuanwumen Bridge | within $2^{nd}$ ring |
| 3 | Xizhimen Bridge | $2^{nd}$ ring road |
| 4 | Fuchenmen | $2^{nd}$ ring road |
| 5 | Baofushi Bridge | $4^{th}$ ring road |
| 6 | Zhongguanchun1 Bridge | $4^{th}$ ring road |
| 7 | Jianguomen | $2^{nd}$ ring road |
| 8 | Madian Bridge | $3^{rd}$ ring road |
| 9 | Sitong Bridge | $3^{rd}$ ring road |
| 10 | Xinwai Street-Dexi Street | $2^{nd}$ ring road |

**Table 2. Important crossroads rank using baseline**

| Rank | Name or Position | Scope |
|---|---|---|
| 1 | Deshenmen bridge | $2^{nd}$ ring road |
| 2 | Xuanwumen bridge | within $2^{nd}$ ring |
| 3 | Beijing Station-Jianwai Street | within $2^{nd}$ ring |
| 4 | Xidan | within $2^{nd}$ ring |
| 5 | Gulouda Street | $2^{nd}$ ring road |
| 6 | Chegongzhuang | $2^{nd}$ ring road |
| 7 | Yuetanbei Bridge | $2^{nd}$ ring road |
| 8 | Dongdan | within $2^{nd}$ ring |
| 9 | Yonghe Palace | within $2^{nd}$ ring |
| 10 | Xizhimen bridge | $2^{nd}$ ring road |

Also, as can be seen, CRRank locates larger number of important crossroads on the $3^{rd}$ and $4^{th}$ ring roads. This is consistent with our daily experience. With the continuous expansion of the city, many residential areas migrate outside $4^{th}$ ring or $5^{th}$ ring, while CBD, shopping and entertainment districts locate near the center of city. There is no doubt that this migration results in the emergence of increasing number of long trips.

Another function of CRRank is that, the tansition load and path popularity can be evaluated. Four tansitions and correspongding main paths are showed in Figure 10, Transition $t_1$ from Jinrong Street to airport has 4607 trips, which is the maximum number in all transtions, CRRank finds that transtion $t_1$ imposes the highest load to road network on account of passing so many important crossroads, such as Xizhimen and Deshengmen etc, and the path of transtion $t_1$ is also the most popular. Transition $t_2$ is from WangJing to Wudaokou, has 495 trips. The initial rank of transition $t_2$ is lower than transition $t_3$, which is from Xuanwumen to Houhai and has 721 trips. But in the result of CRRank, the rank of transition $t_2$ is higher than $t_3$. This is because the trips of $t_2$ traverse larger number of more important crossroads located on $4^{th}$ ring roads. Although the traffic flow of tansition $t_2$ is smaller, a confluence of $t_2$ and other transitions at these important crossroads, may generates a greater burden to road network. Given last example, despite the path of transition $t_4$ from Beijing University of Posts and Telecommunications to Military Museum contains many important crossroads, this tansition has a weak impact on road network, for it has only 7 corresponding trips, and CRRank merely boost rank of transition $t_4$ from $4134^{th}$ to $3475^{th}$.

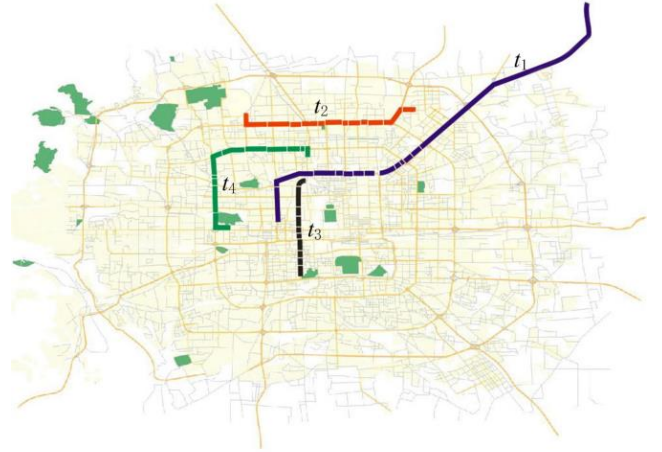

**Figure 10. An example of evaluation of different transitions and paths**

## 6. CONCLUSION
In this paper, we propose an importance of crossroads ranking methods named CRRank based on massive taxi trajectories. Different from previous studies of roles of nodes in road network, CRRank utilizes the dynamic information of road network. A tripartite graph is built to model trip network which combines route knowledge of drivers, travel demands and traffic information. Then an algorithm of score propagation over the tripartite graph is presented to rank the important crossroads. Experiments on trajectories dataset are generated by approximate 10000 taxicabs over one month. The results show that, firstly, compared with the baseline algorithm, CRRank is inclined to discover the important crossroads with high connectivity, especially, the crossroads which are located on the ring roads, are traversed by trips of a wider range of OD pairs, therefore obtain more importance score. This is in accordant with our experiences. Secondly, additional advantages of CRRank are evaluating the load of road network caused by different OD pairs and finding the $k^{th}$ most popular paths. These functions can provide valuable information for transportation planning. In addition, CRRank can effectively discover network-wide important crossroads based on data-driven design.

## 7. ACKNOWLEDGMENTS

Internet of Things", which is under the State High-Tech Development Plan (The 863 program) and funded by The Ministry of Science and Technology of the People's Republic of China (Project No. 2012AA063303).Internet of Things", which is under the State High-Tech Development Plan (The 863 program) and funded by The Ministry of Science and Technology of the People's Republic of China (Project No. 2012AA063303).